\documentclass[letterpaper]{article} 
\usepackage{aaai25}  
\usepackage{times}  
\usepackage{helvet}  
\usepackage{courier}  
\usepackage[hyphens]{url}  
\usepackage{graphicx} 
\usepackage{natbib}  
\usepackage{caption} 
\usepackage{algorithm}
\usepackage{algorithmic}
\usepackage{booktabs} 
\usepackage{multirow} 
\usepackage{graphicx}
\usepackage{array}
\newcolumntype{L}[1]{>{\raggedright\let\newline\\\arraybackslash\hspace{0pt}}m{#1}}
\newcolumntype{C}[1]{>{\centering\let\newline\\\arraybackslash\hspace{0pt}}m{#1}}
\newcolumntype{R}[1]{>{\raggedleft\let\newline\\\arraybackslash\hspace{0pt}}m{#1}}
\usepackage{newfloat}
\usepackage{listings}
\DeclareCaptionStyle{ruled}{labelfont=normalfont,labelsep=colon,strut=off} 
\floatstyle{ruled}
\newfloat{listing}{tb}{lst}{}
\floatname{listing}{Listing}
\usepackage{amsmath}
\title{Debate Helps Weak-to-Strong Generalization}
\author{
    Hao Lang,
    Fei Huang,
    Yongbin Li \thanks{Corresponding author.}
}
\affiliations{
    Tongyi Lab\\
    \{hao.lang, f.huang, shuide.lyb\}@alibaba-inc.com
}

\usepackage{bibentry}

\begin{document}

\maketitle

\begin{abstract}
Common methods for aligning already-capable models with desired behavior rely on the ability of humans to provide supervision.
However, future superhuman models will surpass the capability of humans.
Therefore, humans will only be able to weakly supervise superhuman models.
This expected deficiency of human evaluation would weaken the safety of future AI systems.
Scalable oversight and weak-to-strong generalization are two complementary approaches to tackle this issue.
In this paper, we attempt to combine the strengths of these two approaches to further improve alignment.
Specifically, we investigate ways of improving human supervision with a strong pretrained model and then supervise the strong model with enhanced weak human supervision.
To make iterative empirical progress, we consider an analogy: can we use a strong model to improve weak model supervision and then use it to supervise the strong model?
We empirically test it by finetuning a small weak model on ground truth labels with the additional help from a large strong model, and then finetuning the strong model on labels generated by the weak model.
We find that debate can assist a weak model in extracting trustworthy information from an untrustworthy strong model, which provides leverage as context on samples when training a weak model.
We also show that an ensemble of weak models helps exploit long arguments generated by strong model debaters and obtain a more robust supervision estimate.
Extensive experiments on the OpenAI weak-to-strong NLP benchmarks show that the combination approach leads to better alignment, which indicates that debate has the potential to help weak-to-strong generalization.
\end{abstract}

%

\section{Introduction}

Current AI alignment techniques heavily rely on the availability of human labelled data, such as human demonstrations for supervised finetuning (SFT)~\cite{wei2021finetuned,chung2024scaling} and human preferences for reinforcement learning from human feedback (RLHF)~\cite{christiano2017deep,ouyang2022training,bai2022training}.
These techniques can be leveraged to build the most capable AI systems currently deployed~\cite{OpenAI2023Gpt-4,Anthropic2023Introducing}.

However, as models grow increasingly more capable, they will surpass the ability of humans~\cite{CAIS2023Statement}.
In that case, even human experts can not reliably verify the quality or correctness of model outputs, and the role of human evaluation will evolve into non-experts overseeing experts~\cite{amodei2016concrete,bowman2022measuring,burns2023weak,khan2024debating}.
The expected deficiency of human evaluation will limit the effectiveness of most existing alignment approaches~\cite{casper2023open,mcaleese2024llm}.
Moreover, these predicted inaccurate training signals could lead to reward overoptimization and reward tampering during policy training that seriously weakens its safety~\cite{gao2023scaling,denison2024sycophancy}.

There are two complementary approaches to tackle the above issue: scalable oversight (SO) and weak-to-strong generalization (W2SG)~\cite{JAN2023Combining}.
SO approaches aim to improve the ability of humans to supervise more capable models, such that accurately labelled data can be used for alignment~\cite{bowman2022measuring}.
Instead of improving human supervision, W2SG approaches finetune a strong pretrained model to generalize accurately from weak human supervision~\cite{burns2023weak}.

We note that most prior SO and W2SG techniques are studied separately.
In contrast, we attempt to combine the strength of SO and W2SG to further improve AI alignment.
We investigate ways of improving human supervision with a strong pretrained model and then supervise the strong model with enhanced weak human supervision.
To make iterative empirical progress, we consider an analogy~\cite{burns2023weak,kenton2024scalable}: can we use a strong model to improve weak model supervision and then use it to supervise the strong model?

In this paper, we empirically test it by finetuning a small weak model on ground truth labels with the additional help of knowledge from a large strong model, and then finetuning the strong model on labels generated by the weak model.
We assume a strong model pretrained on internet-scale data can provide contextual information on samples when training a weak model~\cite{brown2020language}.
This gives us hope that a weak-strong model team could create a better weak supervisor to elicit the capabilities of the strong model.

A major challenge in building a weak-strong model team involves finding ways of extracting trustworthy information from untrustworthy models~\cite{bowman2022measuring}.
More specifically, strong pretrained models have huge capabilities but are not well aligned with human values and intentions~\cite{leike2018scalable,ji2023ai}.
Thus, a strong model may intentionally mislead us by generating unfaithful facts or making false claims, which could cause damage when applied for creating a weak supervisor~\cite{michael2023debate}.

Another challenge is that a weak model (with a small model size) may lack the capacity to fully process long contexts generated by the strong model, which are filled with irrelevant noises for tasks at hand.
Meanwhile, recent studies also show that the performance of large language models (LLMs) is closely related to the model size and the complexity of hard problems may exceed the capacity of a single weak model~\cite{xu2023retrieval,chung2024scaling}.

In this study, we demonstrate that debate can help weak models more reliably extract information from strong models.
Concretely, given a question, two instances of a strong pretrained model are randomly assigned two opposing answers, and then the two instances (debaters) argue with each other over the answer~\cite{michael2023debate,khan2024debating,kenton2024scalable}.
In a debate, it is harder to lie than to refute a lie, i.e., if a debater makes false claims, its opponent can convincingly point out flaws in its arguments~\cite{irving2018ai}.
Hence, these arguments from the debate can inform a weak model about the merits and flaws of each side and provide leverage as contextual information in weak model training.

To fully exploit long arguments generated by strong model debaters, we train an ensemble of weak models.
We aggregate predictions of multiple weak models to obtain a more robust supervision estimate over any single one~\cite{ganaie2022ensemble}.
In particular, we explore two types of ensembles: \textit{debate ensembles}, where different members of the ensemble differ in the random seed used during debate sampling, and \textit{finetune ensembles}, where members differ only in the random seed used during weak model finetuning.
We find that debate ensembles consistently outperform a single weak model and finetune ensembles.
The main contributions of this study are summarized as follows:

\begin{itemize}
    \item We show the first demonstration of a simple combination of scalable oversight and weak-to-strong generalization approaches, which leads to better alignment on the OpenAI weak-to-strong NLP benchmarks. 
    \item We find that debate can assist a weak model in extracting trustworthy information from a capable but untrustworthy strong model, which provides leverage as contextual information on samples when training a weak model. We also show that debate outperforms alternative scalable oversight techniques in our settings. 
    \item We show that an ensemble of weak models helps obtain a more robust supervision estimate. We find that diversity of the ensemble is vital, and that a debate ensemble that contains members that do not share a debate sampling seed leads to better performances.
\end{itemize}

Although with the help of knowledge from a strong model, creating a better weak supervisor to elicit the capabilities of the strong model is only one way in which scalable oversight and weak-to-strong generalization techniques can be combined, our results pave the way for further research on hybrid superhuman alignment methods~\cite{JAN2023Combining}.
We provide empirical evidences in NLP domains indicating that debate helps weak-to-strong generalization.

\begin{figure*}[t]
\centering
\includegraphics[width = 0.83\linewidth]{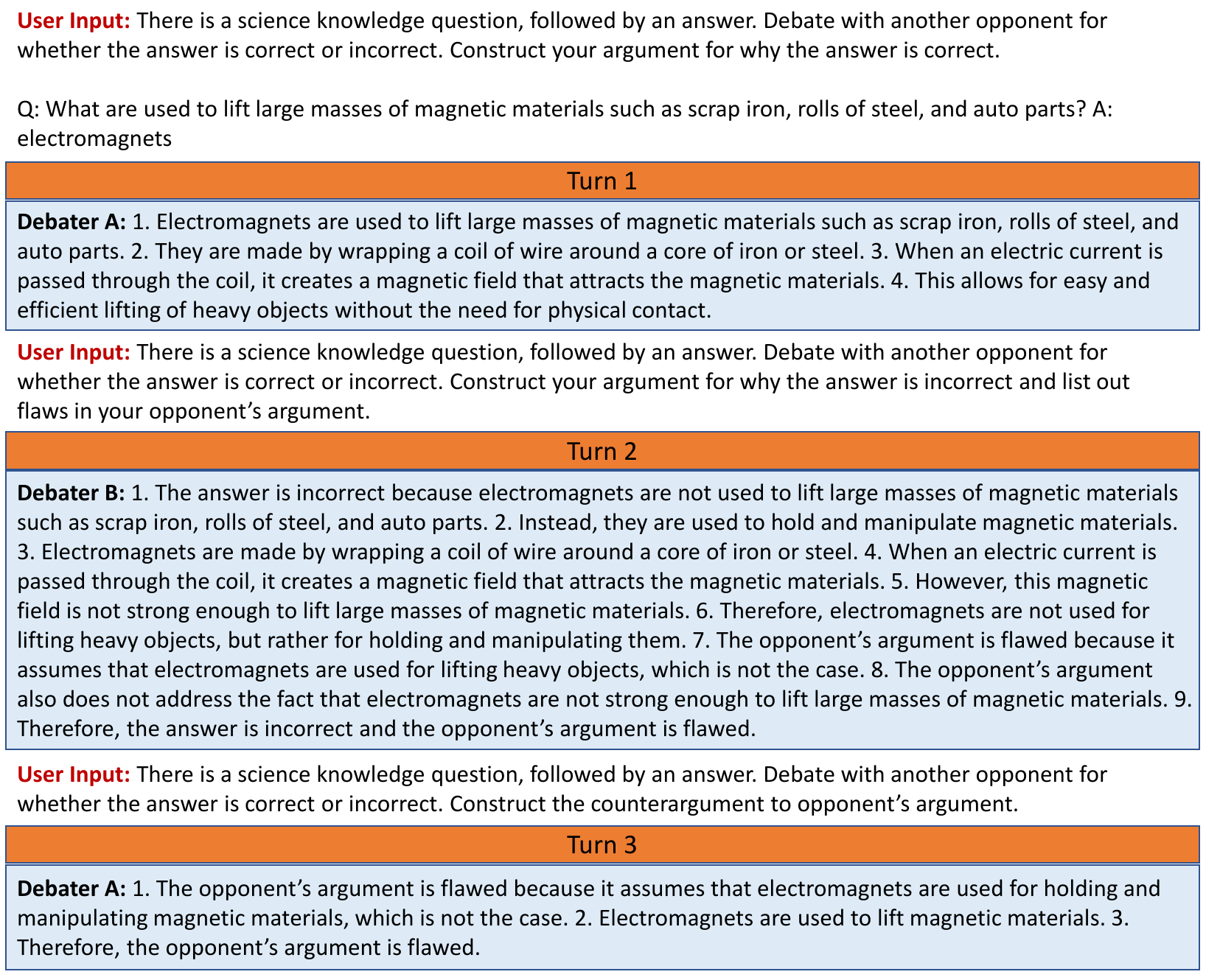}
\caption{\textbf{Illustration of debate}. Illustration of the debate procedure between debater A and debater B.}
\label{fig:framework}
\end{figure*}

\section{Related Work}

\textbf{AI alignment.} The goal of AI alignment is to steer already-capable models to behave in line with human values and intentions~\cite{leike2018scalable,ji2023ai}.
Current alignment methods finetune pretrained LLMs using imitation learning on human demonstrations~\cite{bain1995framework,atkeson1997robot,wei2021finetuned,chung2024scaling}, reinforcement learning from human feedback (RLHF)~\cite{christiano2017deep,stiennon2020learning,ouyang2022training,bai2022training}, or direct alignment algorithms like direct preference optimization (DPO)~\cite{rafailov2024direct,rafailov2024scaling}.
Both imitation learning and preference learning rely on high-quality human supervision, a demand that becomes increasingly challenging as models become more capable than humans~\cite{amodei2016concrete}.

\paragraph{Scalable oversight.} Scalable oversight techniques seek to improve the ability of humans to supervise more capable models~\cite{bowman2022measuring}.
This is typically pursued through taking advantage of special problem structure, such as the assumption that evaluation is easier than generation~\cite{karp1975computational,goodfellow2014generative} or decomposability~\cite{christiano2018supervising}.
There have been many promising scalable oversight proposals in theory, including Recursive Reward Modeling~\cite{leike2018scalable}, Debate~\cite{irving2018ai}, Market-Making~\cite{Hubinger2020market}, Self-Critique~\cite{saunders2022self}, and many more~\cite{lightman2023let,mcaleese2024llm,sun2024easy}.
Recent empirical studies in this direction demonstrate that human-machine teams can improve evaluation accuracy on question answering tasks over the human-only baseline~\cite{bowman2022measuring}.

Debate was originally proposed for AI safety~\cite{irving2018ai}.
From then on, a body of work has explored the usability of debate for scalable oversight, with human or LLM debaters~\cite{parrish2022single,parrish2022two,michael2023debate,khan2024debating,kenton2024scalable}.
These studies are all conducted to improve inference-time judge accuracy, while in our work debate is leveraged to train a better weak supervisor.
We could in turn use the weak supervisor to align strong models.
LLM-based debate has also been investigated in several other applications, like translation~\cite{liang2023encouraging}, text assessment~\cite{chan2023chateval}, reasoning and content generation~\cite{du2023improving}. 

\paragraph{Weak-to-strong generalization.} In contrast to improving human supervision, weak-to-strong generalization techniques finetune a strong pretrained model to generalize well from weak human supervision~\cite{burns2023weak}.
The hope for these techniques is that strong pretrained models should already have good representations of the alignment-relevant tasks.
Therefore, we simply need a weak supervisor to elicit what the strong model already knows.
Recently, a theoretical framework is introduced to understand weak-to-strong generalization with misfit error~\cite{charikar2024quantifying}.
Prior work has mainly explored how to supervise a strong model with a fixed weak supervisor, while in this work we also attempt to train a better weak supervisor with the help of the strong model.

\paragraph{Ensemble methods.} Our work is also related to existing works that use ensembles by combining predictions of several models~\cite{ganaie2022ensemble}.
In the context of AI alignment, reward model ensembles are investigated to mitigate reward overoptimization when finetuning models with RLHF~\cite{coste2023reward,eisenstein2023helping}.
Most similar to our work, \citet{liu2024co} propose to assemble a diverse set of specialized weak supervisors for weak-to-strong generalization.
In contrast, in our work, we aim to use an ensemble of weak models with different seeds to fully exploit long arguments generated by strong model debaters.

\section{Preliminaries}

We review the weak-to-strong generalization pipeline in~\cite{burns2023weak}, which has also been adopted in subsequent work~\cite{liu2024co,charikar2024quantifying}.
It usually consists of three phases:

\paragraph{1. Create the weak supervisor.} We create the weak supervisor by finetuning a small pretrained model on ground truth labels.
We call the performance of the weak supervisor the \textit{weak performance}.

\paragraph{2. Train a strong student model.} We train a strong student model by finetuning a large pretrained model on weak labels generated by the weak supervisor.
We call its performance the \textit{weak-to-strong performance}.

\paragraph{3. Train a strong ceiling model.} We train a strong ceiling model by finetuning a large pretrained model on ground truth labels.
We call this model’s resulting performance the \textit{strong ceiling performance}.

To measure the fraction of the performance gap that the strong student model can recover with weak supervision, we define the performance gap recovered (PGR) using the above three performances:

\begin{equation} 
\mathrm{PGR}= \frac{\text{weak-to-strong}-\text{weak}}{\text{strong ceiling}-\text{weak}}. \nonumber
\end{equation}

\begin{table*}[t]
    \centering
    \small
    \begin{tabular}{C{50pt} |C{35pt} |  L{360pt} }
    \toprule
    \textbf{Debate Turn} & \textbf{Debater} & \textbf{Prompt} \\
    \midrule
     First & A & \textit{There is a science knowledge question, followed by an answer}. Debate with another opponent for whether the answer is \textit{correct} or \textit{incorrect}. Construct your argument for why the answer is \textit{correct}.\\
    \midrule
     Second & B & \textit{There is a science knowledge question, followed by an answer}. Debate with another opponent for whether the answer is \textit{correct} or \textit{incorrect}. Construct your argument for why the answer is \textit{incorrect} and list out flaws in your opponent’s argument.\\
     \midrule
     Third & A & \textit{There is a science knowledge question, followed by an answer}. Debate with another opponent for whether the answer is \textit{correct} or \textit{incorrect}. Construct the counterargument to opponent’s argument.\\
    \bottomrule
    \end{tabular}
    \caption{\textbf{Prompts to induce debate on a binary classification problem}. The binary classification problem is converted from the SciQ dataset. Two answer choices \textit{correct} and \textit{incorrect} are randomly assigned to debater A and B. Debate runs for three turns. We also append the current debate transcript to the prompt.}
    \label{tab:prompt}
\end{table*}

\section{Methods}

\subsection{Overview}

In this study, we build the strong student model following three steps:
\textbf{1.} Generate arguments from the debate between two instances of a large pretrained model;
\textbf{2.} Train an ensemble of weak models using these debate arguments;
\textbf{3.} Train a strong student model using labels that are constructed by the weak model ensemble.

\subsection{Argument Generation through Debate}

We assume large pretrained models embed broad-coverage knowledge that can help a variety of tasks~\cite{brown2020language}.
Our goal is to extract trustworthy information from a capable but untrustworthy strong model via debate~\cite{bowman2022measuring}.
So we could in turn use the trustworthy information to help train a better weak model.

We first describe the debate protocol we investigated to elicit truth from strong models, following~\cite{michael2023debate,khan2024debating,kenton2024scalable}.
Given a question and its two answer choices (one correct, one incorrect), two instances of a large pretrained model (debaters) are randomly assigned to argue for these two opposing answers.
Debate is turn-based textual exchanges between the two debaters, which take turns to review arguments from previous turns and generate their arguments for the next turn.
After a pre-determined number of turns, the debate is ended and the transcript of arguments from the debate is kept.
During the debate, each debater presents the most compelling evidences for its assigned answer and arguments to explain why its opponent’s claims are false.

Concretely, debate runs for three turns in this work.
At the start of a turn, debaters are prompted with instructions outlining the problem, their assigned answer, and the current debate transcript.
The prompts to induce debate are illustrated in Table~\ref{tab:prompt}.

We illustrate an overview of this debate procedure in Figure~\ref{fig:framework}.
We can observe that Debater B is on the side of an incorrect answer and incentivized to present misleading arguments.
However, in the next turn, Debater A convincingly points out these false claims and thus Debater B can not easily get away.
This observation conforms to the claim, i.e., it is harder to lie than to refute a lie.~\cite{irving2018ai}.
These arguments from the debate can provide valuable information about the merits and flaws of each side, which have the potential to significantly advance the capabilities of weak models.

\subsection{Weak Model Ensemble Training}

For each input sample of weak models, we append it with the kept debate transcript.
We train a weak model by finetuning a small pretrained model on these augmented samples with ground truth labels.
We note that the debate transcripts generated in a multi-turn debate are long, which may be difficult for a weak model to fully process.
Therefore, we train an ensemble of weak models $\{W_1,...,W_k\}$ to help improve robustness~\cite{lakshminarayanan2017simple}.

We explore two types of ensembles: \textit{debate ensembles}, where the debate transcript used by each member is generated with a different random seed, and \textit{finetune ensembles}, where all members share the same debate transcript, but use a different seed when finetuned on the augmented samples.
Debate ensembles are much more expensive to train, but are more diverse and thus likely to lead to a more robust prediction.
Unless stated otherwise, we train an ensemble consisting of four individual weak models.

\subsection{Training Strong Models using Ensembles}

We finally train a strong student model by finetuning a large pretrained model on weak labels constructed by the weak model ensemble.
We simply take the mean of the predictions from different weak models within the ensemble as the weak label for each training sample~\cite{ganaie2022ensemble}.

\section{Experiments}

\subsection{Tasks}

We adopt the evaluation protocol of prior work~\cite{burns2023weak}, and conduct experiments in NLP tasks on four classification datasets: SciQ~\cite{welbl2017crowdsourcing}, BoolQ~\cite{clark2019boolq}, CosmosQA~\cite{huang2019cosmos}, and AnthropicHH~\cite{bai2022training}.
We convert each dataset to a binary classification problem.
For multiple-choice datasets, given a data point with a question $Q$ and $k$ candidate answers $A$, we construct $k$ new data points of the form $(Q, A_i)$, where the label is $1$ for the correct
answers and $0$ for all the incorrect answers.
We also keep the same number of correct and incorrect answers per question to maintain class balance.

\begin{table*}[t]
    \centering
    \small
    
    \begin{tabular}{l|l|cc|cc|cc|cc}
    \toprule
    \multicolumn{1}{l|}{\multirow{2}{*}{\textbf{Performance}}} & \multicolumn{1}{l|}{\multirow{2}{*}{\textbf{Method}}} & \multicolumn{2}{c|}{\textbf{SciQ}} & \multicolumn{2}{c|}{\textbf{BoolQ}} & \multicolumn{2}{c|}{\textbf{CosmosQA}} & \multicolumn{2}{c}{\textbf{AnthropicHH}} \\
    \multicolumn{1}{l|}{} & \multicolumn{1}{l|}{} &  Acc. &  PGR &  Acc. &  PGR &  Acc. &  PGR &  Acc. &  PGR \\
    \midrule
    Weak performance & & 90.0 &   & 86.0 &  & 87.5 &  & 48.8 &  \\
    \midrule
    \multirow{4}{*}{Weak-to-strong performance} & Finetune & 91.5 & 44.1 & 88.0 & 51.3& 88.2 & 30.4& 49.5 & 35.0\\
    & Finetune w/ aux. loss & 91.4 & 41.2 & 88.2 & 56.4& 87.9 & 17.4& 49.5 & 35.0\\
    & Finetune w/ pro. loss & 91.6 & 47.1 & 88.1 & 53.8& 88.1 & 26.1&  49.2 & 20.0\\
    & \textbf{Ours} & \textbf{92.6} & \textbf{76.5} & \textbf{88.7} & \textbf{69.2} & \textbf{88.8} & \textbf{56.5}& \textbf{50.2} & \textbf{70.0}\\
    \midrule
    Strong ceiling performance & & 93.4 &   & 89.9 &  & 89.8 &  & 50.8 &  \\
    \bottomrule
    \end{tabular}
    \caption{\textbf{Debate improves weak-to-strong generalization}. Test accuarcy (\%) and performance gap recovered (PGR) (\%) of our approach and baselines on the binary classification tasks converted from NLP classification datasets. Here, our approach uses debate ensembles. Accuracy of weak and strong models trained with ground truth are reported as weak performance and strong ceiling performance, respectively. }
    \label{tab:main_result}
\end{table*}

\subsection{Experimental Setups and Metrics}

We randomly sample at most 20k data points from each task and split them in half.
We train a weak model on the first half of the data points and use its prediction on the other half as the weak labels.
The weak labels are soft labels~\cite{hinton2015distilling}.
We report the accuracy and performance gap recovered (PGR) of the strong student model on the test set in all tasks.
The weak performance for PGR is the performance of the naively finetuned small model.

\subsection{Implementation Details}

Our implementations of data preprocessing, weak and strong model training are based on the OpenAI weak-to-strong codebase and its default hyper-parameters~\cite{burns2023weak}.
Specifically, we use Qwen/Qwen-7B as the small pretrained model for training weak models.
Meanwhile, we use Qwen/Qwen-14B as the large pretrained model for generating debate arguments and training strong models.
Both Qwen/Qwen-7B and Qwen/Qwen-14B are open-sources, which can aid reproducibility~\cite{bai2023qwen}.
We do not use pretrained models from the GPT-2 family for training weak models~\cite{radford2019language}, because they lack the capability required for scalable oversight techniques like working closely with strong models~\cite{bowman2022measuring}.

For each converted binary classification problem, we use the two candidate answers per question as the two opposing answers, which are randomly assigned to the two strong model debaters in a debate.
In order to adapt weak and strong models to the binary classification setting, we equip each model with a linear classification head with two outputs on top of the encoder.
We train all models for two epochs with a batch size of 32.
We conduct all experiments on a single 8×A100 machine.

\subsection{Baselines}

We compare our approach with competitive baseline approaches:
1. \textbf{Finetune}~\cite{burns2023weak} naively finetunes strong pretrained models on labels generated by a weak model;
2. \textbf{Finetune w/ aux. loss}~\cite{burns2023weak} finetunes strong models with an auxiliary confidence loss, which reinforces the strong model’s confidence in its own predictions when they disagree with the weak labels;
3. \textbf{Finetune w/ pro. loss}~\cite{burns2023weak} finetunes strong models with a confidence-like loss which sets the cross entropy targets to the product of weak labels and strong model predictions.
We also report the \textbf{weak performance} and the \textbf{strong ceiling performance} defined in the preliminaries section.
Note that the strong ceiling performance is generally regarded as the upper bound of the \textbf{weak-to-strong performance} when only weak labels are considered.

\subsection{Main Results}

In Table~\ref{tab:main_result}, we report the results of each approach on the binary classification tasks converted from SciQ, BoolQ, CosmosQA, and AnthropicHH datasets.
Here, our approach uses debate ensembles.
In each task, we observe that PGRs of strong student models finetuned on weak labels are all positive.
This indicates that student models consistently outperform their weak supervisors across all weak-to-strong generation approaches and tasks that we studied.
Simultaneously, this promising weak-to-strong generalization also suggests that our experimental settings can help make iterative empirical progress in tackling the weak supervision issue for aligning future superhuman models.

At the same time, we find that our approach significantly outperforms each strong student baseline, including the naive baseline finetuned on weak labels or more sophisticated baselines equipped with a confidence loss term on all four tasks.
Compared with the promising baseline Finetune w/ aux. loss, our approach brings up from a PGR of $41.2\%$ to $76.5\%$ in SciQ, $56.4\%$ to $69.2\%$ in BoolQ, $17.4\%$ to $56.5\%$ in CosmosQA, and $35.0\%$ to $70.0\%$ in AnthropicHH.
Our approach also obtains the best test accuracy among all compared strong students.
The performance gain demonstrates the advantage of extracting trustworthy information from the strong model via debate, which helps create a better weak supervisor to elicit the capabilities of the strong model.

In addition, we also see that adding a confidence loss to the standard cross entropy objective (Finetune w/ aux. loss and Finetune w/ pro. loss) generally gives a modest boost in generalization performance.
In our experimental settings, the gaps in compute between weak and strong models are not significantly large, which may limit their performances.

\subsection{Ablation Studies}

\begin{table*}[t]
    \centering
    \small
    \begin{tabular}{L{60pt} |  L{360pt} }
    \toprule
    \textbf{Protocol} & \textbf{Prompt} \\
    \midrule
     Consultancy   & \textit{There is a science knowledge question, followed by an answer}. Construct your argument for why the answer is \textit{[random answer]}.\\
    \midrule
     Market-Making & \textit{There is a science knowledge question, followed by an answer}. Construct your argument for why the answer is \textit{[unselected answer]}.\\
    \bottomrule
    \end{tabular}
    \caption{\textbf{Prompts to induce consultancy and market-making}. The binary classification problem is converted from the SciQ dataset. The two answer choices are correct and incorrect. \textit{[random answer]} is the answer randomly sampled from the two candidate answers. \textit{[unselected answer]} is the answer that is not selected by the weak supervisor (the naively finetuned small model) based on its prediction. We also append the current transcript to the prompt.}
    \label{tab:prompt2}
\end{table*}

\begin{table*}[t]
    \centering
    \small
    \begin{tabular}{l|cc|cc|cc|cc}
    \toprule
    \multicolumn{1}{l|}{\multirow{2}{*}{\textbf{Method}}} & \multicolumn{2}{c|}{\textbf{SciQ}} & \multicolumn{2}{c|}{\textbf{BoolQ}} & \multicolumn{2}{c|}{\textbf{CosmosQA}} & \multicolumn{2}{c}{\textbf{AnthropicHH}} \\
    \multicolumn{1}{l|}{} &  Acc. &  PGR &  Acc. &  PGR &  Acc. &  PGR &  Acc. &  PGR \\
    \midrule
    Consultancy & 91.5 & 44.1 & 87.8 & 46.2 & 88.3 & 34.8& 49.3 & 25.0\\
    Market-Making & 91.6 & 47.1 & 87.6 & 41.0& 88.2 & 30.4& 49.5 & 35.0\\
    \midrule
    \textbf{Ours} & \textbf{92.6} & \textbf{76.5} & \textbf{88.7} & \textbf{69.2} & \textbf{88.8} & \textbf{56.5}& \textbf{50.2} & \textbf{70.0}\\
    \bottomrule
    \end{tabular}
   \caption{\textbf{Ablation on different scalable oversight approaches}. Here, our approach uses debate ensembles.}
    \label{tab:ab_so}
\end{table*}

Finally, we provide comprehensive ablation studies to understand the efficacy of debate for weak-to-strong generation.

\paragraph{Ablation on different scalable oversight approaches.} We demonstrate the effectiveness of debate as a mechanism to extract trustworthy information from a capable but untrustworthy strong model by replacing it with other alternative scalable oversight approaches: Consultancy~\cite{michael2023debate} and Market-Making~\cite{Hubinger2020market}.
\begin{itemize}
    \item \textbf{Consultancy.} In consultancy, there is only one consultant instead of two debaters. The consultant is an instance of a large pretrained model. Given a question and its two answer choices (one correct, one incorrect), the consultant is assigned to argue for one of these answers, with a 50\% chance of each. During the consultancy, the consultant provides evidences for its assigned answer. The transcript is kept at the end of the consultancy.
    \item \textbf{Market-Making}. In market-making, there is a single debater. The debater is an instance of a large pretrained model and aims to generate arguments that change some model's beliefs on the answer to a question. Given a question and its two candidate answers, we let the weak supervisor (the naively finetuned small model) select an answer based on its prediction. Accordingly, the debater is assigned to argue for the unselected answer. The transcript is kept at the end.
\end{itemize}

Specifically, consultancy and market-making run for a single turn. At the start of each turn, the consultant and the debater are provided with a prompt describing the task, the assigned answer, and the transcript.
The prompts to induce consultancy and market-making are illustrated in Table~\ref{tab:prompt2}.

Results in Table~\ref{tab:ab_so} show that debate used in our approach performs better than all other variants in terms of test accuracy and PGR across all four tasks.
These results validate our claim that debate can help elicit truth from a strong model, at least better than consultancy and market-making in our settings.
Meanwhile, we should note that consultancy is a relatively weak baseline to beat, because there is a 50-50 chance of the consultant arguing for the incorrect answer.

\begin{table*}[h]
    \centering
    \small
    \begin{tabular}{l|cc|cc|cc|cc}
    \toprule
    \multicolumn{1}{l|}{\multirow{2}{*}{\textbf{Method}}} & \multicolumn{2}{c|}{\textbf{SciQ}} & \multicolumn{2}{c|}{\textbf{BoolQ}} & \multicolumn{2}{c|}{\textbf{CosmosQA}} & \multicolumn{2}{c}{\textbf{AnthropicHH}} \\
    \multicolumn{1}{l|}{} &  Acc. &  PGR &  Acc. &  PGR &  Acc. &  PGR &  Acc. &  PGR \\
    \midrule
    Single weak model & 91.7 & 50.0 & 88.2 & 56.4 & 88.4 & 39.1& 49.5 & 35.0\\
    Finetune ensembles & 91.8 & 52.9 & 88.3 & 59.0& 88.4 & 39.1& 49.7 & 45.0\\
    \midrule
    \textbf{debate ensembles} & \textbf{92.6} & \textbf{76.5} & \textbf{88.7} & \textbf{69.2} & \textbf{88.8} & \textbf{56.5}& \textbf{50.2} & \textbf{70.0}\\
    \bottomrule
    \end{tabular}
   \caption{Ablation on weak model ensemble. }
    \label{tab:ab_em}
\end{table*}

\begin{figure*}[t]
\centering
\includegraphics[width = 0.55\linewidth]{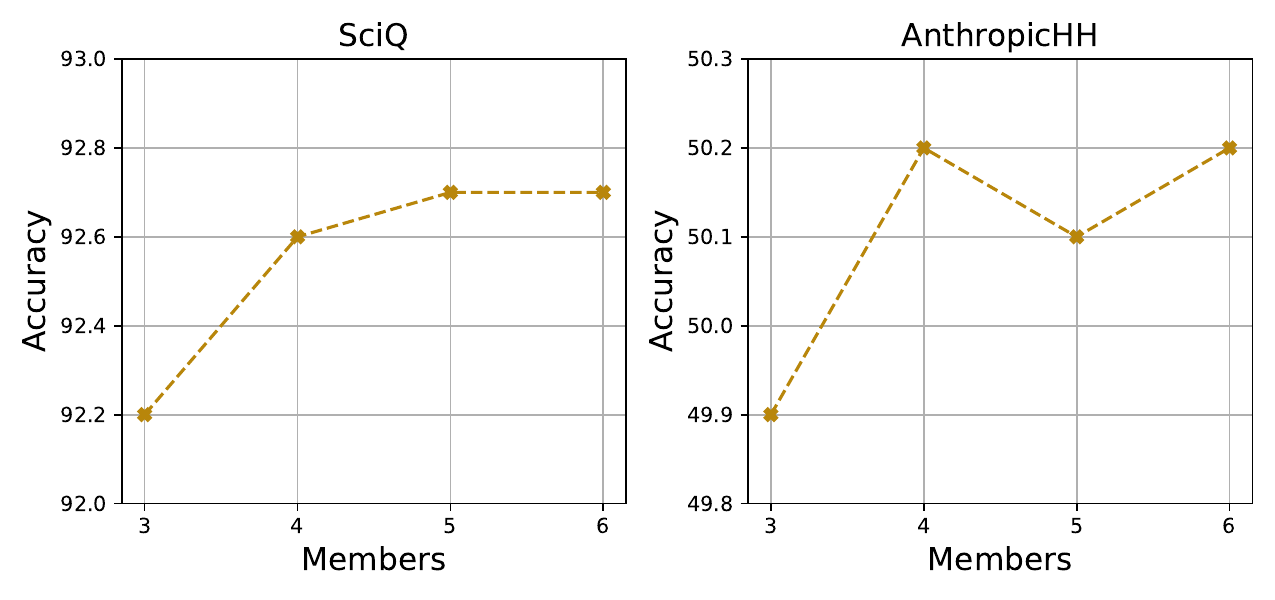}
\caption{\textbf{Ablation study on the cardinality of the ensemble}. Here, our approach uses debate ensembles.}
\label{fig:cardinality}
\end{figure*}

\begin{figure*}[!h]
\centering
\includegraphics[width = 0.55\linewidth]{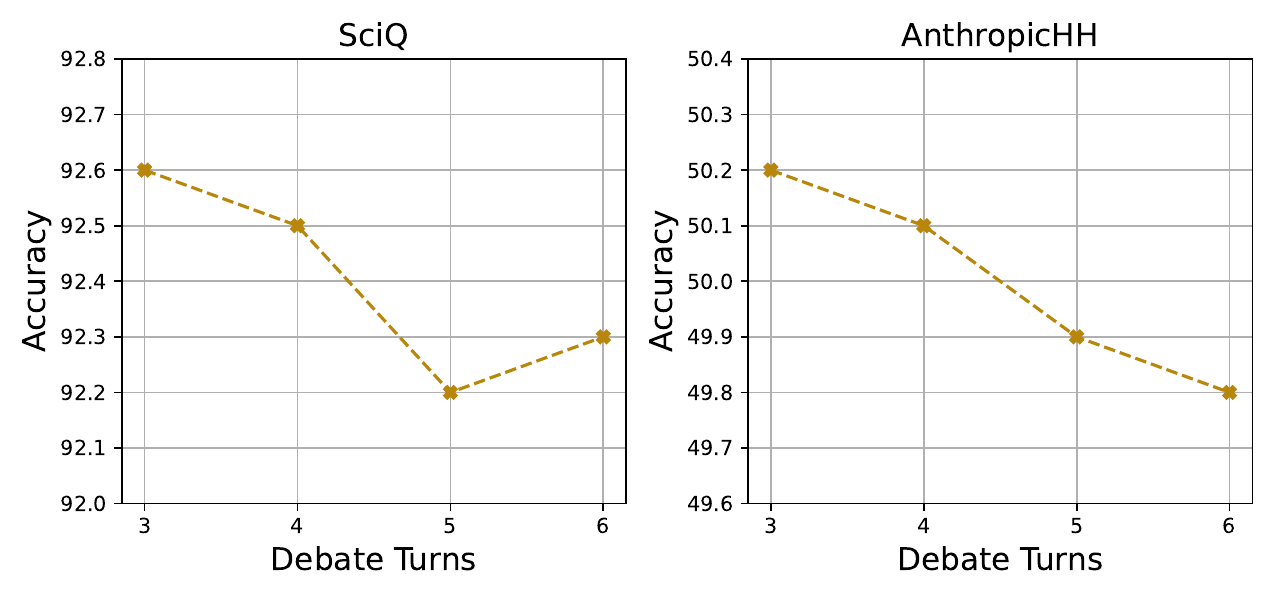}
\caption{\textbf{Ablation on the number of turns of debate}. Here, our approach uses debate ensembles.}
\label{fig:turns}
\end{figure*}

\paragraph{Ablation on weak model ensemble.} We analyze the impact of weak model ensemble on obtaining a robust weak supervision estimate for weak-to-strong generalization.
In Table~\ref{tab:ab_em}, we compare three weak model ensemble methods of increasing computational cost: single weak model, finetune ensembles, and debate ensembles.
Single weak model is an individual small model finetuned on the samples augmented with debate transcripts.
Finetune ensembles and debate ensembles are described in the methods section.

We find that debate ensembles consistently improve performance over individual weak models and finetune ensembles across all the tasks.
On the contrary, finetune ensembles relatively improve performance over individual weak models in 3 out of 4 tasks and are comparable in the other.
These results suggest that the diversity of generated debate arguments is the key to the success of weak model ensemble, which helps create a better weak supervisor.
At the same time, an individual small weak model may lack the capability to fully exploit long arguments from the debate, as a result, leading to limited performances.

\paragraph{Ablation on the cardinality of the ensemble.} Recall that our weak model ensemble method introduces an additional hyperparameter cardinality.
The cardinality is the size of the ensemble.
We analyze the impact of the cardinality of the ensemble on the final performance.
In Figure~\ref{fig:cardinality}, we increase weak model members used in the ensemble on SciQ and AnthropicHH tasks.
We can observe that there is a significant gap between 3-member and 4-member ensembles.
On the other hand, the performance of 4-member, 5-member, and 6-member ensembles is quite similar.
It suggests that 4-member ensemble is likely to work best and diminishing returns will occur after this point.

\paragraph{Ablation on the number of turns of debate.} Next, we analyze the impact of the number of turns of debate on the final performance.
In Figure~\ref{fig:turns}, we increase the debate length for up to 6 turns on SciQ and AnthropicHH tasks.
We find that more turns of debate do not increase the final performance.
We observe that strong model debaters like Qwen/Qwen-14B suffer from the inability to effectively process long debate transcripts and follow instructions, as turns continue, as shown by a consistent decrease in test accuracy after 3 turns.
We used 3 turns of debate in this work because it is the minimum interaction between debaters to extract the truth from the strong model.
For instance, the two debaters can critique their opponent in turn 2 and turn 3, respectively.

\section{Limitations and Conclusion}

\paragraph{Limitations.}
In this work, we attempt to combine the strength of two complementary approaches, i.e., scalable oversight and weak-to-strong generalization, to tackle the issue of weak supervision for aligning future superhuman models.
For this purpose, we explore a simple combination method, i.e., extracting trustworthy information via debate from strong models and using it to create a better weak supervisor to elicit the capabilities of strong models.
Although our proposed method is found to be effective in all our experiments and ablation studies, there are many more ways to combine scalable oversight and weak-to-strong generalization, such as Task decomposition + W2SG~\cite{JAN2023Combining}.
More empirical work is needed in this area.

In our setup, the difference between strong and weak models is only in the size of pretrained models. In the future, stronger models may also differ in reasoning and planning abilities.
Furthermore, the gaps in compute between weak and strong models are not significantly large in this work (7B vs. 14B).
It would be interesting to verify our conclusions on more large and advanced models, such as Qwen/Qwen2-72B~\cite{qwen2}.
Finally, our approach is expensive as it requires both two instances of debaters and a multi-turn debate procedure.

\paragraph{Conclusion.}
In this paper, we present an approach to improve the performance of weak-to-strong generalization via debate.
We believe the perspective of having scalable oversight and weak-to-strong generalization methods working in combination to tackle the weak supervision issue will prove to be a fruitful area of research in superhuman alignment.

\bibliography{aaai25}

\begin{thebibliography}{54}
\providecommand{\natexlab}[1]{#1}

\bibitem[{Amodei et~al.(2016)Amodei, Olah, Steinhardt, Christiano, Schulman, and Man{\'e}}]{amodei2016concrete}
Amodei, D.; Olah, C.; Steinhardt, J.; Christiano, P.; Schulman, J.; and Man{\'e}, D. 2016.
\newblock Concrete problems in AI safety.
\newblock \emph{arXiv preprint arXiv:1606.06565}.

\bibitem[{Anthropic(2023)}]{Anthropic2023Introducing}
Anthropic. 2023.
\newblock Introducing claude.
\newblock \url{https://www.anthropic.com/index/introducing-claude}.

\bibitem[{Atkeson and Schaal(1997)}]{atkeson1997robot}
Atkeson, C.~G.; and Schaal, S. 1997.
\newblock Robot learning from demonstration.
\newblock In \emph{ICML}, volume~97, 12--20.

\bibitem[{Bai et~al.(2023)Bai, Bai, Chu, Cui, Dang, Deng, Fan, Ge, Han, Huang et~al.}]{bai2023qwen}
Bai, J.; Bai, S.; Chu, Y.; Cui, Z.; Dang, K.; Deng, X.; Fan, Y.; Ge, W.; Han, Y.; Huang, F.; et~al. 2023.
\newblock Qwen technical report.
\newblock \emph{arXiv preprint arXiv:2309.16609}.

\bibitem[{Bai et~al.(2022)Bai, Jones, Ndousse, Askell, Chen, DasSarma, Drain, Fort, Ganguli, Henighan et~al.}]{bai2022training}
Bai, Y.; Jones, A.; Ndousse, K.; Askell, A.; Chen, A.; DasSarma, N.; Drain, D.; Fort, S.; Ganguli, D.; Henighan, T.; et~al. 2022.
\newblock Training a helpful and harmless assistant with reinforcement learning from human feedback.
\newblock \emph{arXiv preprint arXiv:2204.05862}.

\bibitem[{Bain and Sammut(1995)}]{bain1995framework}
Bain, M.; and Sammut, C. 1995.
\newblock A Framework for Behavioural Cloning.
\newblock In \emph{Machine Intelligence 15}, 103--129.

\bibitem[{Bowman et~al.(2022)Bowman, Hyun, Perez, Chen, Pettit, Heiner, Luko{\v{s}}i{\=u}t{\.e}, Askell, Jones, Chen et~al.}]{bowman2022measuring}
Bowman, S.~R.; Hyun, J.; Perez, E.; Chen, E.; Pettit, C.; Heiner, S.; Luko{\v{s}}i{\=u}t{\.e}, K.; Askell, A.; Jones, A.; Chen, A.; et~al. 2022.
\newblock Measuring progress on scalable oversight for large language models.
\newblock \emph{arXiv preprint arXiv:2211.03540}.

\bibitem[{Brown et~al.(2020)Brown, Mann, Ryder, Subbiah, Kaplan, Dhariwal, Neelakantan, Shyam, Sastry, Askell et~al.}]{brown2020language}
Brown, T.; Mann, B.; Ryder, N.; Subbiah, M.; Kaplan, J.~D.; Dhariwal, P.; Neelakantan, A.; Shyam, P.; Sastry, G.; Askell, A.; et~al. 2020.
\newblock Language models are few-shot learners.
\newblock \emph{Advances in neural information processing systems}, 33: 1877--1901.

\bibitem[{Burns et~al.(2023)Burns, Izmailov, Kirchner, Baker, Gao, Aschenbrenner, Chen, Ecoffet, Joglekar, Leike et~al.}]{burns2023weak}
Burns, C.; Izmailov, P.; Kirchner, J.~H.; Baker, B.; Gao, L.; Aschenbrenner, L.; Chen, Y.; Ecoffet, A.; Joglekar, M.; Leike, J.; et~al. 2023.
\newblock Weak-to-strong generalization: Eliciting strong capabilities with weak supervision.
\newblock \emph{arXiv preprint arXiv:2312.09390}.

\bibitem[{CAIS(2023)}]{CAIS2023Statement}
CAIS. 2023.
\newblock Statement on AI Risk.
\newblock \url{https://www.safe.ai/work/statement-on-ai-risk}.

\bibitem[{Casper et~al.(2023)Casper, Davies, Shi, Gilbert, Scheurer, Rando, Freedman, Korbak, Lindner, Freire et~al.}]{casper2023open}
Casper, S.; Davies, X.; Shi, C.; Gilbert, T.~K.; Scheurer, J.; Rando, J.; Freedman, R.; Korbak, T.; Lindner, D.; Freire, P.; et~al. 2023.
\newblock Open problems and fundamental limitations of reinforcement learning from human feedback.
\newblock \emph{arXiv preprint arXiv:2307.15217}.

\bibitem[{Chan et~al.(2023)Chan, Chen, Su, Yu, Xue, Zhang, Fu, and Liu}]{chan2023chateval}
Chan, C.-M.; Chen, W.; Su, Y.; Yu, J.; Xue, W.; Zhang, S.; Fu, J.; and Liu, Z. 2023.
\newblock Chateval: Towards better llm-based evaluators through multi-agent debate.
\newblock \emph{arXiv preprint arXiv:2308.07201}.

\bibitem[{Charikar, Pabbaraju, and Shiragur(2024)}]{charikar2024quantifying}
Charikar, M.; Pabbaraju, C.; and Shiragur, K. 2024.
\newblock Quantifying the Gain in Weak-to-Strong Generalization.
\newblock \emph{arXiv preprint arXiv:2405.15116}.

\bibitem[{Christiano, Shlegeris, and Amodei(2018)}]{christiano2018supervising}
Christiano, P.; Shlegeris, B.; and Amodei, D. 2018.
\newblock Supervising strong learners by amplifying weak experts.
\newblock \emph{arXiv preprint arXiv:1810.08575}.

\bibitem[{Christiano et~al.(2017)Christiano, Leike, Brown, Martic, Legg, and Amodei}]{christiano2017deep}
Christiano, P.~F.; Leike, J.; Brown, T.; Martic, M.; Legg, S.; and Amodei, D. 2017.
\newblock Deep reinforcement learning from human preferences.
\newblock \emph{Advances in neural information processing systems}, 30.

\bibitem[{Chung et~al.(2024)Chung, Hou, Longpre, Zoph, Tay, Fedus, Li, Wang, Dehghani, Brahma et~al.}]{chung2024scaling}
Chung, H.~W.; Hou, L.; Longpre, S.; Zoph, B.; Tay, Y.; Fedus, W.; Li, Y.; Wang, X.; Dehghani, M.; Brahma, S.; et~al. 2024.
\newblock Scaling instruction-finetuned language models.
\newblock \emph{Journal of Machine Learning Research}, 25(70): 1--53.

\bibitem[{Clark et~al.(2019)Clark, Lee, Chang, Kwiatkowski, Collins, and Toutanova}]{clark2019boolq}
Clark, C.; Lee, K.; Chang, M.-W.; Kwiatkowski, T.; Collins, M.; and Toutanova, K. 2019.
\newblock BoolQ: Exploring the surprising difficulty of natural yes/no questions.
\newblock \emph{arXiv preprint arXiv:1905.10044}.

\bibitem[{Coste et~al.(2023)Coste, Anwar, Kirk, and Krueger}]{coste2023reward}
Coste, T.; Anwar, U.; Kirk, R.; and Krueger, D. 2023.
\newblock Reward model ensembles help mitigate overoptimization.
\newblock \emph{arXiv preprint arXiv:2310.02743}.

\bibitem[{Denison et~al.(2024)Denison, MacDiarmid, Barez, Duvenaud, Kravec, Marks, Schiefer, Soklaski, Tamkin, Kaplan et~al.}]{denison2024sycophancy}
Denison, C.; MacDiarmid, M.; Barez, F.; Duvenaud, D.; Kravec, S.; Marks, S.; Schiefer, N.; Soklaski, R.; Tamkin, A.; Kaplan, J.; et~al. 2024.
\newblock Sycophancy to Subterfuge: Investigating Reward-Tampering in Large Language Models.
\newblock \emph{arXiv preprint arXiv:2406.10162}.

\bibitem[{Du et~al.(2023)Du, Li, Torralba, Tenenbaum, and Mordatch}]{du2023improving}
Du, Y.; Li, S.; Torralba, A.; Tenenbaum, J.~B.; and Mordatch, I. 2023.
\newblock Improving factuality and reasoning in language models through multiagent debate.
\newblock \emph{arXiv preprint arXiv:2305.14325}.

\bibitem[{Eisenstein et~al.(2023)Eisenstein, Nagpal, Agarwal, Beirami, D'Amour, Dvijotham, Fisch, Heller, Pfohl, Ramachandran et~al.}]{eisenstein2023helping}
Eisenstein, J.; Nagpal, C.; Agarwal, A.; Beirami, A.; D'Amour, A.; Dvijotham, D.; Fisch, A.; Heller, K.; Pfohl, S.; Ramachandran, D.; et~al. 2023.
\newblock Helping or herding? reward model ensembles mitigate but do not eliminate reward hacking.
\newblock \emph{arXiv preprint arXiv:2312.09244}.

\bibitem[{Ganaie et~al.(2022)Ganaie, Hu, Malik, Tanveer, and Suganthan}]{ganaie2022ensemble}
Ganaie, M.~A.; Hu, M.; Malik, A.~K.; Tanveer, M.; and Suganthan, P.~N. 2022.
\newblock Ensemble deep learning: A review.
\newblock \emph{Engineering Applications of Artificial Intelligence}, 115: 105151.

\bibitem[{Gao, Schulman, and Hilton(2023)}]{gao2023scaling}
Gao, L.; Schulman, J.; and Hilton, J. 2023.
\newblock Scaling laws for reward model overoptimization.
\newblock In \emph{International Conference on Machine Learning}, 10835--10866. PMLR.

\bibitem[{Goodfellow et~al.(2014)Goodfellow, Pouget-Abadie, Mirza, Xu, Warde-Farley, Ozair, Courville, and Bengio}]{goodfellow2014generative}
Goodfellow, I.; Pouget-Abadie, J.; Mirza, M.; Xu, B.; Warde-Farley, D.; Ozair, S.; Courville, A.; and Bengio, Y. 2014.
\newblock Generative adversarial nets.
\newblock \emph{Advances in neural information processing systems}, 27.

\bibitem[{Hinton, Vinyals, and Dean(2015)}]{hinton2015distilling}
Hinton, G.; Vinyals, O.; and Dean, J. 2015.
\newblock Distilling the knowledge in a neural network.
\newblock \emph{arXiv preprint arXiv:1503.02531}.

\bibitem[{Huang et~al.(2019)Huang, Bras, Bhagavatula, and Choi}]{huang2019cosmos}
Huang, L.; Bras, R.~L.; Bhagavatula, C.; and Choi, Y. 2019.
\newblock Cosmos QA: Machine reading comprehension with contextual commonsense reasoning.
\newblock \emph{arXiv preprint arXiv:1909.00277}.

\bibitem[{Hubinger(2020)}]{Hubinger2020market}
Hubinger, E. 2020.
\newblock AI safety via market making.
\newblock \emph{AI Alignment Forum}.

\bibitem[{Irving, Christiano, and Amodei(2018)}]{irving2018ai}
Irving, G.; Christiano, P.; and Amodei, D. 2018.
\newblock AI safety via debate.
\newblock \emph{arXiv preprint arXiv:1805.00899}.

\bibitem[{Ji et~al.(2023)Ji, Qiu, Chen, Zhang, Lou, Wang, Duan, He, Zhou, Zhang et~al.}]{ji2023ai}
Ji, J.; Qiu, T.; Chen, B.; Zhang, B.; Lou, H.; Wang, K.; Duan, Y.; He, Z.; Zhou, J.; Zhang, Z.; et~al. 2023.
\newblock Ai alignment: A comprehensive survey.
\newblock \emph{arXiv preprint arXiv:2310.19852}.

\bibitem[{Karp(1975)}]{karp1975computational}
Karp, R.~M. 1975.
\newblock On the computational complexity of combinatorial problems.
\newblock \emph{Networks}, 5(1): 45--68.

\bibitem[{Kenton et~al.(2024)Kenton, Siegel, Kram{\'a}r, Brown-Cohen, Albanie, Bulian, Agarwal, Lindner, Tang, Goodman et~al.}]{kenton2024scalable}
Kenton, Z.; Siegel, N.~Y.; Kram{\'a}r, J.; Brown-Cohen, J.; Albanie, S.; Bulian, J.; Agarwal, R.; Lindner, D.; Tang, Y.; Goodman, N.~D.; et~al. 2024.
\newblock On scalable oversight with weak LLMs judging strong LLMs.
\newblock \emph{arXiv preprint arXiv:2407.04622}.

\bibitem[{Khan et~al.(2024)Khan, Hughes, Valentine, Ruis, Sachan, Radhakrishnan, Grefenstette, Bowman, Rockt{\"a}schel, and Perez}]{khan2024debating}
Khan, A.; Hughes, J.; Valentine, D.; Ruis, L.; Sachan, K.; Radhakrishnan, A.; Grefenstette, E.; Bowman, S.~R.; Rockt{\"a}schel, T.; and Perez, E. 2024.
\newblock Debating with more persuasive llms leads to more truthful answers.
\newblock \emph{arXiv preprint arXiv:2402.06782}.

\bibitem[{Lakshminarayanan, Pritzel, and Blundell(2017)}]{lakshminarayanan2017simple}
Lakshminarayanan, B.; Pritzel, A.; and Blundell, C. 2017.
\newblock Simple and scalable predictive uncertainty estimation using deep ensembles.
\newblock \emph{Advances in neural information processing systems}, 30.

\bibitem[{Leike(2023)}]{JAN2023Combining}
Leike, J. 2023.
\newblock Combining weak-to-strong generalization with scalable oversight.

\bibitem[{Leike et~al.(2018)Leike, Krueger, Everitt, Martic, Maini, and Legg}]{leike2018scalable}
Leike, J.; Krueger, D.; Everitt, T.; Martic, M.; Maini, V.; and Legg, S. 2018.
\newblock Scalable agent alignment via reward modeling: a research direction.
\newblock \emph{arXiv preprint arXiv:1811.07871}.

\bibitem[{Liang et~al.(2023)Liang, He, Jiao, Wang, Wang, Wang, Yang, Tu, and Shi}]{liang2023encouraging}
Liang, T.; He, Z.; Jiao, W.; Wang, X.; Wang, Y.; Wang, R.; Yang, Y.; Tu, Z.; and Shi, S. 2023.
\newblock Encouraging divergent thinking in large language models through multi-agent debate.
\newblock \emph{arXiv preprint arXiv:2305.19118}.

\bibitem[{Lightman et~al.(2023)Lightman, Kosaraju, Burda, Edwards, Baker, Lee, Leike, Schulman, Sutskever, and Cobbe}]{lightman2023let}
Lightman, H.; Kosaraju, V.; Burda, Y.; Edwards, H.; Baker, B.; Lee, T.; Leike, J.; Schulman, J.; Sutskever, I.; and Cobbe, K. 2023.
\newblock Let's verify step by step.
\newblock \emph{arXiv preprint arXiv:2305.20050}.

\bibitem[{Liu and Alahi(2024)}]{liu2024co}
Liu, Y.; and Alahi, A. 2024.
\newblock Co-supervised learning: Improving weak-to-strong generalization with hierarchical mixture of experts.
\newblock \emph{arXiv preprint arXiv:2402.15505}.

\bibitem[{McAleese et~al.(2024)McAleese, Pokorny, Uribe, Nitishinskaya, Trebacz, and Leike}]{mcaleese2024llm}
McAleese, N.; Pokorny, R.~M.; Uribe, J. F.~C.; Nitishinskaya, E.; Trebacz, M.; and Leike, J. 2024.
\newblock LLM Critics Help Catch LLM Bugs.
\newblock \emph{arXiv preprint arXiv:2407.00215}.

\bibitem[{Michael et~al.(2023)Michael, Mahdi, Rein, Petty, Dirani, Padmakumar, and Bowman}]{michael2023debate}
Michael, J.; Mahdi, S.; Rein, D.; Petty, J.; Dirani, J.; Padmakumar, V.; and Bowman, S.~R. 2023.
\newblock Debate helps supervise unreliable experts.
\newblock \emph{arXiv preprint arXiv:2311.08702}.

\bibitem[{OpenAI(2023)}]{OpenAI2023Gpt-4}
OpenAI. 2023.
\newblock Gpt-4 technical report.
\newblock \url{https://openai.com/index/gpt-4-research/}.

\bibitem[{Ouyang et~al.(2022)Ouyang, Wu, Jiang, Almeida, Wainwright, Mishkin, Zhang, Agarwal, Slama, Ray et~al.}]{ouyang2022training}
Ouyang, L.; Wu, J.; Jiang, X.; Almeida, D.; Wainwright, C.; Mishkin, P.; Zhang, C.; Agarwal, S.; Slama, K.; Ray, A.; et~al. 2022.
\newblock Training language models to follow instructions with human feedback.
\newblock \emph{Advances in neural information processing systems}, 35: 27730--27744.

\bibitem[{Parrish et~al.(2022{\natexlab{a}})Parrish, Trivedi, Nangia, Padmakumar, Phang, Saimbhi, and Bowman}]{parrish2022two}
Parrish, A.; Trivedi, H.; Nangia, N.; Padmakumar, V.; Phang, J.; Saimbhi, A.~S.; and Bowman, S.~R. 2022{\natexlab{a}}.
\newblock Two-Turn Debate Doesn't Help Humans Answer Hard Reading Comprehension Questions.
\newblock \emph{arXiv preprint arXiv:2210.10860}.

\bibitem[{Parrish et~al.(2022{\natexlab{b}})Parrish, Trivedi, Perez, Chen, Nangia, Phang, and Bowman}]{parrish2022single}
Parrish, A.; Trivedi, H.; Perez, E.; Chen, A.; Nangia, N.; Phang, J.; and Bowman, S.~R. 2022{\natexlab{b}}.
\newblock Single-turn debate does not help humans answer hard reading-comprehension questions.
\newblock \emph{arXiv preprint arXiv:2204.05212}.

\bibitem[{Radford et~al.(2019)Radford, Wu, Child, Luan, Amodei, Sutskever et~al.}]{radford2019language}
Radford, A.; Wu, J.; Child, R.; Luan, D.; Amodei, D.; Sutskever, I.; et~al. 2019.
\newblock Language models are unsupervised multitask learners.
\newblock \emph{OpenAI blog}, 1(8): 9.

\bibitem[{Rafailov et~al.(2024{\natexlab{a}})Rafailov, Chittepu, Park, Sikchi, Hejna, Knox, Finn, and Niekum}]{rafailov2024scaling}
Rafailov, R.; Chittepu, Y.; Park, R.; Sikchi, H.; Hejna, J.; Knox, B.; Finn, C.; and Niekum, S. 2024{\natexlab{a}}.
\newblock Scaling laws for reward model overoptimization in direct alignment algorithms.
\newblock \emph{arXiv preprint arXiv:2406.02900}.

\bibitem[{Rafailov et~al.(2024{\natexlab{b}})Rafailov, Sharma, Mitchell, Manning, Ermon, and Finn}]{rafailov2024direct}
Rafailov, R.; Sharma, A.; Mitchell, E.; Manning, C.~D.; Ermon, S.; and Finn, C. 2024{\natexlab{b}}.
\newblock Direct preference optimization: Your language model is secretly a reward model.
\newblock \emph{Advances in Neural Information Processing Systems}, 36.

\bibitem[{Saunders et~al.(2022)Saunders, Yeh, Wu, Bills, Ouyang, Ward, and Leike}]{saunders2022self}
Saunders, W.; Yeh, C.; Wu, J.; Bills, S.; Ouyang, L.; Ward, J.; and Leike, J. 2022.
\newblock Self-critiquing models for assisting human evaluators.
\newblock \emph{arXiv preprint arXiv:2206.05802}.

\bibitem[{Stiennon et~al.(2020)Stiennon, Ouyang, Wu, Ziegler, Lowe, Voss, Radford, Amodei, and Christiano}]{stiennon2020learning}
Stiennon, N.; Ouyang, L.; Wu, J.; Ziegler, D.; Lowe, R.; Voss, C.; Radford, A.; Amodei, D.; and Christiano, P.~F. 2020.
\newblock Learning to summarize with human feedback.
\newblock \emph{Advances in Neural Information Processing Systems}, 33: 3008--3021.

\bibitem[{Sun et~al.(2024)Sun, Yu, Shen, Liu, Yang, Welleck, and Gan}]{sun2024easy}
Sun, Z.; Yu, L.; Shen, Y.; Liu, W.; Yang, Y.; Welleck, S.; and Gan, C. 2024.
\newblock Easy-to-hard generalization: Scalable alignment beyond human supervision.
\newblock \emph{arXiv preprint arXiv:2403.09472}.

\bibitem[{Wei et~al.(2021)Wei, Bosma, Zhao, Guu, Yu, Lester, Du, Dai, and Le}]{wei2021finetuned}
Wei, J.; Bosma, M.; Zhao, V.~Y.; Guu, K.; Yu, A.~W.; Lester, B.; Du, N.; Dai, A.~M.; and Le, Q.~V. 2021.
\newblock Finetuned language models are zero-shot learners.
\newblock \emph{arXiv preprint arXiv:2109.01652}.

\bibitem[{Welbl, Liu, and Gardner(2017)}]{welbl2017crowdsourcing}
Welbl, J.; Liu, N.~F.; and Gardner, M. 2017.
\newblock Crowdsourcing multiple choice science questions.
\newblock \emph{arXiv preprint arXiv:1707.06209}.

\bibitem[{Xu et~al.(2023)Xu, Ping, Wu, McAfee, Zhu, Liu, Subramanian, Bakhturina, Shoeybi, and Catanzaro}]{xu2023retrieval}
Xu, P.; Ping, W.; Wu, X.; McAfee, L.; Zhu, C.; Liu, Z.; Subramanian, S.; Bakhturina, E.; Shoeybi, M.; and Catanzaro, B. 2023.
\newblock Retrieval meets long context large language models.
\newblock \emph{arXiv preprint arXiv:2310.03025}.

\bibitem[{Yang et~al.(2024)Yang, Yang, Hui, Zheng, Yu, Zhou, Li, Li, Liu, Huang, Dong, Wei, Lin, Tang, Wang, Yang, Tu, Zhang, Ma, Xu, Zhou, Bai, He, Lin, Dang, Lu, Chen, Yang, Li, Xue, Ni, Zhang, Wang, Peng, Men, Gao, Lin, Wang, Bai, Tan, Zhu, Li, Liu, Ge, Deng, Zhou, Ren, Zhang, Wei, Ren, Fan, Yao, Zhang, Wan, Chu, Liu, Cui, Zhang, and Fan}]{qwen2}
Yang, A.; Yang, B.; Hui, B.; Zheng, B.; Yu, B.; Zhou, C.; Li, C.; Li, C.; Liu, D.; Huang, F.; Dong, G.; Wei, H.; Lin, H.; Tang, J.; Wang, J.; Yang, J.; Tu, J.; Zhang, J.; Ma, J.; Xu, J.; Zhou, J.; Bai, J.; He, J.; Lin, J.; Dang, K.; Lu, K.; Chen, K.; Yang, K.; Li, M.; Xue, M.; Ni, N.; Zhang, P.; Wang, P.; Peng, R.; Men, R.; Gao, R.; Lin, R.; Wang, S.; Bai, S.; Tan, S.; Zhu, T.; Li, T.; Liu, T.; Ge, W.; Deng, X.; Zhou, X.; Ren, X.; Zhang, X.; Wei, X.; Ren, X.; Fan, Y.; Yao, Y.; Zhang, Y.; Wan, Y.; Chu, Y.; Liu, Y.; Cui, Z.; Zhang, Z.; and Fan, Z. 2024.
\newblock Qwen2 Technical Report.
\newblock \emph{arXiv preprint arXiv:2407.10671}.

\end{thebibliography}

\end{document}